\documentclass[10pt,conference]{IEEEtran}
\IEEEoverridecommandlockouts
\makeatletter
\def\ps@headings{%
\def\@oddhead{\mbox{}\scriptsize\rightmark \hfil \thepage}%
\def\@evenhead{\scriptsize\thepage \hfil \leftmark\mbox{}}%
\def\@oddfoot{}%
\def\@evenfoot{}}
\makeatother
\usepackage{cite}
\usepackage[numbers]{natbib}
\usepackage{amsmath,amssymb,amsfonts}
\usepackage{graphicx}
\usepackage{textcomp}
\usepackage{xcolor}
\def\BibTeX{{\rm B\kern-.05em{\sc i\kern-.025em b}\kern-.08em
    T\kern-.1667em\lower.7ex\hbox{E}\kern-.125emX}}
\usepackage[margin=0.625in,top=0.75in,bottom=1in]{geometry}
\usepackage{multirow}
\usepackage{balance}
\usepackage{amsmath}
\usepackage{amsfonts}
\usepackage{comment}
\usepackage{algorithm}
\usepackage{algpseudocode}
\def\NoNumber#1{{\def\alglinenumber##1{}\State #1}\addtocounter{ALG@line}{-1}}

\begin{document}

\title{Adversarial Attacks Neutralization via Data Set Randomization\\
}
\author{\IEEEauthorblockN{Mouna Rabhi}
\IEEEauthorblockA{\textit{Division of Information and Computing Technology} \\
\textit{College of Science and Engineering}\\
\textit{Hamad Bin Khalifa University}\\
Doha, Qatar \\
mora33056@hbku.edu.qa}
\and
\IEEEauthorblockN{Roberto Di Pietro}
\IEEEauthorblockA{\textit{Resilient Computing and Cybersecurity Center (RC3)} \\
\textit{Division of Computer, Electrical and Mathematical }\\
\textit{Science  and Engineering (CEMSE)}\\
\textit{King Abdullah University of Science and Technology}\\
roberto.dipietro@kaust.edu.sa}
}

\maketitle

\begin{abstract}

Adversarial attacks on deep neural models pose a serious threat to their reliability  and security. Existing defense mechanisms are often narrow in focus, addressing a specific type of attack, or being vulnerable to sophisticated attacks.
 In this paper, we propose a new defense mechanism that, while being focused on image-based classifiers,  is general with respect to the cited category. It is rooted on hyperspace projection. In particular, our solution provides a pseudo-random projection of the original dataset into a new dataset.
The proposed defense mechanism creates a set of diverse projected datasets, where  each projected dataset is used to train a specific classifier, resulting in different trained classifiers with different decision boundaries. At every test step, the defense mechanism uses a randomly selected classifier to test the input. What is more, the proposed approach does not sacrifice accuracy over legitimate input.\\
Other than detailing and  providing a thorough characterization of our defense mechanism, we also provide a proof of concept of using four optimization-based adversarial attacks, namely PGD, FGSM, IGSM, and C\&W,  and a generative adversarial attack testing them on the MNIST 
 dataset. Our experimental results show that our solution 
 increases the robustness of deep learning models against adversarial attacks and significantly reduces the attack success rate by at least 89\% for optimization-based attacks and 78\% for generative attacks.  We also analyze the relationship between the number of  used hyperspaces and the efficacy of the defense mechanism. As expected, the two are positively correlated, hence offering an easy-to-tune parameter to enforce the desired level of security. \\*
The generality and scalability of our defense mechanism---not specific to a particular type of attack and easily adaptable to different attack scenarios---combined with the excellent achieved results, other than offering an actionable defense layer against adversarial attacks on deep neural networks, also lay the groundwork for future research in the field. 
\end{abstract}


\section{Introduction}
Deep Learning algorithms have been acknowledged as a significant milestone in the field of machine learning. These algorithms have shown tremendous success in numerous fields and are among the most widely used techniques for pattern recognition, auto-driving, image processing, classification tasks, and more. Moreover, deep learning has found widespread applications in sensitive domains such as finance, health, and military \cite{wang2009brief}.
Notwithstanding the remarkable performance achieved by deep learning models, they are susceptible to adversarial attacks that cause them to make erroneous predictions. Adversarial examples are input samples that have been intentionally altered to elicit a specific response from a model, typically resulting in misclassification or a desired incorrect prediction that would benefit the attacker \cite{akhtar2021advances}. This vulnerability poses a significant threat to the reliability and security of deep learning models. The adversarial examples are crafted with the specific intent of deceiving the model and causing it to produce incorrect predictions, which can have far-reaching consequences in sensitive applications such as healthcare, finance, and autonomous driving.  
For example, researchers have shown that adding imperceptible perturbations to a stop sign  image acquired by an autonomous car can cause the car's object detection system into failing to recognize the sign as a stop sign and misclassify it with high confidence, even though the perturbed image is visually indistinguishable from the original \cite{kang2020adversarial}. Similarly, by adding small amounts of noise to the input of a speech recognition model, an adversary can cause the model to recognize a different word or phrase than what was actually spoken \cite{carlini2018audio}.
Moreover, the creation of adversarial examples requires minimal resources and expertise, making them an accessible tool for attackers to exploit vulnerabilities in machine learning models. Adversarial attacks can also evade traditional security measures, making it challenging to detect and prevent them. As a result, the development of effective countermeasures against adversarial examples is critical for ensuring the robustness and trustworthiness of deep learning models in real-world applications.

Researchers have proposed numerous defense solutions to defend against adversarial attacks, including adversarial training, gradient masking, distillation techniques, input transformation, and adversarial example detection \cite{ shaham2015understanding, papernot2016distillation, nowroozi2022resisting,aprilpyone2021block, li2017adversarial}. However, many of these defenses have been shown to be ineffective against stronger attacks and only serve to mitigate or obscure adversarial weaknesses. Adversarial training, for example, has been found to be prone to overfitting to the specific attack used during training, and it may not be robust to new, previously unseen attacks \cite{yu2022understanding}. Adversarial example detection methods have also been found to be vulnerable to attacks by adversaries \cite{carlini2017adversarial}.

A common solution in the field is to resort to input randomization, which involves adding noise or perturbations to the input data during the training and testing phases of the deep learning model. Input randomization  applies different types of noise, distortion, or other perturbations to the input data before feeding it to the model. This makes it more difficult for attackers to identify specific features or patterns in the input data that can be exploited to generate effective adversarial examples.
This solution is, for instance, used in some recent work \cite{taran2019defending, aprilpyone2021block}. However, while valid, it has been shown to be limited against adaptive attacks \cite{ali2022evaluating}, where  the attacker has gained some knowledge about the targeted model and  uses this knowledge to generate adversarial examples that are tailored to the model's vulnerabilities. As a result, there is a growing need to develop adversarial defense techniques that can maintain high classification accuracy while ensuring robustness against adversarial attacks.

To overcome the cited limitations, we propose a new defense mechanism  to mitigate adversarial attacks. Our solution leverages  hyperspace projection of the input using  random  images. The hyperspace projection is achieved using randomly generated images that transform the model's input from the original hyperspace to another hyperspace.
Our main goal is to design a transformation that can protect the deep neural networks  classifier from adaptive adversarial attacks while preserving the features that allow the model to classify the input with high accuracy. In a nutshell, our solution works as follows:  we generate a set of $r$ random images that will be used to transform the input dataset into $r$ randomly projected datasets. The proposed transformation is used as a preprocessing technique to transform the images during the training and testing phase. We train $r$ different classifiers. At every testing step, we randomly select a random image from the $r$ available random images and retrieve its trained classifier from the $r$ available trained classifiers. The proposed approach introduces randomness, making it difficult for adversarial attacks to generate examples that can fool all the classifiers. Moreover, it  adds an unpredictable element to the input data making it challenging for the attacker to predict or guess the transformed data.\\
The proposed defense mechanism was tested on the MNIST dataset and has shown to be effective against optimization-based as well as generative adversarial attacks. 
The state-of-the-art adversarial optimization-based attacks do not reach an attack success rate above 10\%, which is the random classification rate for the MNIST classifier. The generative adversarial attacks  as well were not able to surpass the random classification rate when the defense mechanism employs a  number of random images just above 4. Moreover, the proposed defense mechanism  reduces the attack success rate by at least $89.7\%$~for optimization-based attacks and $78.97\%$ for generative adversarial attacks.
Finally, we studied the impact of the number of random images used on the robustness of the proposed defense.

{\bf Contributions.} The main contributions of this work can be summarized as follows:
\begin{itemize}
    \item We propose a new defense mechanism that leverages random images for hyperspace projection.    
    \item The proposed defense mechanism is able to  improve the robustness of a classifier against adversarial attacks while maintaining high classification accuracy on legitimate data.
    \item The proposed defense mechanism does not require adding any additional defense neural networks. It can be easily implemented on existing classifiers and requires minimal computational resources, making it a practical and effective solution for improving the security of machine learning models against adversarial attacks.
    \item The proposed solution is fully tunable. As such, it can withstand any devised adversary model.
    \item We tested our solution against state-of-the-art attacks, showing its striking efficacy and viability.     
\end{itemize}
To the best of our knowledge, our work is the first to explore hyperspace projection using random images to defend against adversarial attacks. Based on the initial results demonstrating the effectiveness of our proposed defense mechanism against known attacks, we have confidence that utilizing random images for input projection represents a promising direction for further research in enhancing the security of deep learning models against adversarial attacks.

{\bf Roadmap.} The sequel of this paper is organized as follows: Section \ref{sec:related} overviews the adversarial attacks and defenses proposed in the literature. Section \ref{sec:threat} describes the threat model. Section \ref{sec:dtl_sol} provides a detailed description of our proposed defense mechanism to mitigate adversarial attacks. Section \ref{sec:exp} introduces our experimental methodology and discusses the results of our evaluation. Finally, Section \ref{sec:conclusion} concludes our work.

\section{Related work}
\label{sec:related}
In this section, we will first present the proposed algorithms for generating adversarial attacks. We will then discuss the proposed defenses that have been developed to mitigate adversarial attacks. Table \ref{Tab:acr} provides the list of acronyms and abbreviations used in this paper.
\begin{table}[htbp] 
\centering
\begin{tabular}{cc} \\ \hline 
{\em Acronym} & {\em Description} \\\hline 
  $X$      &   Original dataset         \\ \hline 
  x &      Size of the original dataset      \\ \hline 
  $X_i$    &    An image of the dataset X      \\ \hline 
  R & Set of random images  \\ \hline 
  r & Size of the random image set  \\ \hline 
  $R_j$ & An image from the set R  \\ \hline 
  $X'$ & Set of projected datasets \\ \hline
   $X'_j$ & Projected dataset   \\ \hline 
   $X'_{j,i}$ & An image from the projected dataset \\ \hline 
   In & Input image\\ \hline
   l & Classification label \\ \hline
   f & Classifier  \\ \hline 
   t & target class\\ \hline 
   ASR      &   Attack Success Rate         \\ \hline
   FGSM & FAST Gradient Sign Method \\ \hline
   IGSM & Iterative Gradient Sign Method \\ \hline
   PGD & Projected Gradient Sign \\ \hline
   C\&W & Carlini and Wagner\\ \hline
   PRNG & Pseudo-random Number Generator\\ \hline

\end{tabular}
\caption{Acronym and abbreviation list.}
\label{Tab:acr}
\end{table}
\subsection{Existing adversarial attacks}
Adversarial attacks have been a popular research topic in the field of deep learning, and numerous methods have been proposed to generate adversarial examples that can deceive deep learning models. Adversarial attacks can be categorized into: (1) optimization-based attacks; and, (2) generative attacks. Optimization-based attacks involve finding small perturbations to the input data that can cause the model to misclassify the input. These attacks typically use gradient-based optimization algorithms to iteratively search for the optimal perturbation. Generative attacks involve generating new data that is similar to the original input but has been modified to cause misclassification. These attacks often use generative models like generative adversarial networks (GANs) or variational autoencoders (VAEs) to generate the modified data. Adversarial attacks can be further classified into: (1) adaptive attacks and (2) non-adaptive attacks. Adaptive adversarial attacks often involve iterative process in which the attacker interacts with the target model to generate adversarial examples. The attacker feeds the input to the model, observes the target model's response, and then updates the perturbation based on the target model's behavior. On the other hand, in non-adaptive attacks, the attacker does not have direct access to the target model and has to devise an attack strategy based on prior knowledge or assumptions about the model's structure or behavior. These attacks are typically more limited in their scope and require a certain level of premeditation \\
This section reviews some of the most relevant works on adversarial attacks that are commonly used to test the defense mechanism.\\
\subsubsection{Optimization-based adversarial attacks}
\hfill\\
\textbf{\em{FGSM attack}} The fast gradient sign method is a one-step adversarial attack \cite{goodfellow2014explaining}. Given an image $X_i$, FGSM aims to create an adversarial image $X'_i$, that looks similar to $X$ and is able to fool the classifier. To do so, it computes an adversarial perturbation in the direction of the gradient ($\nabla$) of the model's cost function, $loss(X,l_X)$ that describes the cost of classifying $X_i$ as label $l$, and then  multiplies it by a small value $\epsilon$. The adversarial image $X'_i$ is expressed as follows:
\begin{equation}
    X'_i=X_i+\epsilon. sign(\nabla_{X_i} loss(X_i,l_{X_i}))
\end{equation}
\\
\textbf{\em{IGSM attack}} The iterative gradient sign method is an iterative variant of FGSM proposed by \cite{kurakin2018adversarial}. The IGSM iteratively   perturbs the input image by adding a fixed perturbation value $\epsilon$ multiplied by the sign of the gradient of the loss function with respect to the input. This process is repeated for a certain number of iterations to generate the final adversarial example. The update process at the $i^{th}$ iteration is expressed as follows:
\begin{equation}
    X'_{i+1}=clip_{\epsilon,X_i}(X'_i+\alpha sign(\nabla_{X_i} loss(X_i,l_{X_i})))
\end{equation}
The clip function is used to ensure that the perturbation added to the input image during an attack stays within the $\epsilon$-neighborhood of the input image $x$.
\\
\textbf{\em{PGD attack}} The projected gradient descent was proposed by \cite{kurakin2018adversarial} to improve the FGSM attack. PGD is an iterative adversarial attack that aims to generate adversarial examples by perturbing the input image with small and imperceptible modifications that can fool the model. To do so, it computes the first adversarial image$X_0$, which can be obtained by adding a small perturbation $\delta_0$ to the original input image $X_i$. The initial perturbation $\delta_0$ can be either  zero or random noise. Then the adversarial image $X'_i$ is updated using the following equation:\\
\begin{equation}
    X'_{i+1}=Clip_{(X_i,\epsilon)} \{X_i+\alpha sign(\nabla_{X_i} loss({X_i},l_{X_i}))\}
\end{equation}
where $=Clip_{(X_i,\epsilon)}^z=min\{255,X_i+\epsilon, max\{0,X_i-\epsilon,z\}\}$ is a function that clips the adversarial example to ensure that it stays within the $\epsilon$ 
 neighborhood around the original input.  $alpha$ is the step size or learning rate of the attack, and $\epsilon$ is the maximum Lp-norm bound of the perturbation.
 \\
\textbf{\em{ C\&W attack}} The Carlini and Wagner attack is regarded as one of the most powerful adversarial attacks \cite{carlini2017towards}. C\&W attack is able to generate adversarial examples that are highly effective and visually indistinguishable from the original inputs. The attack can be untargeted or targeted  and can employ three distance metrics: $L_0$, $L_2$, and $L_\infty$. In $L_2$ attack, given an input image $X_i$, a target class $t$ different from the original class of $X_i$, the attack optimization problem can be expressed as follows: 
\begin{equation}
    minimize ||X'_i-X_i||_2^2+ c * loss(X'_i, t)
\end{equation}
where $c$ is a hyperparameter that controls the tradeoff between the distance of the adversarial example from the original image and the effectiveness of the attack, and $loss(X'_i, t)$ is a loss function that measures the distance between the predicted class probabilities for the adversarial example $X'_i$ and the target class $t$.
\subsubsection{Generative adversarial attacks}
\begin{figure}[htbp]
\centerline{\includegraphics[width=0.9\linewidth]{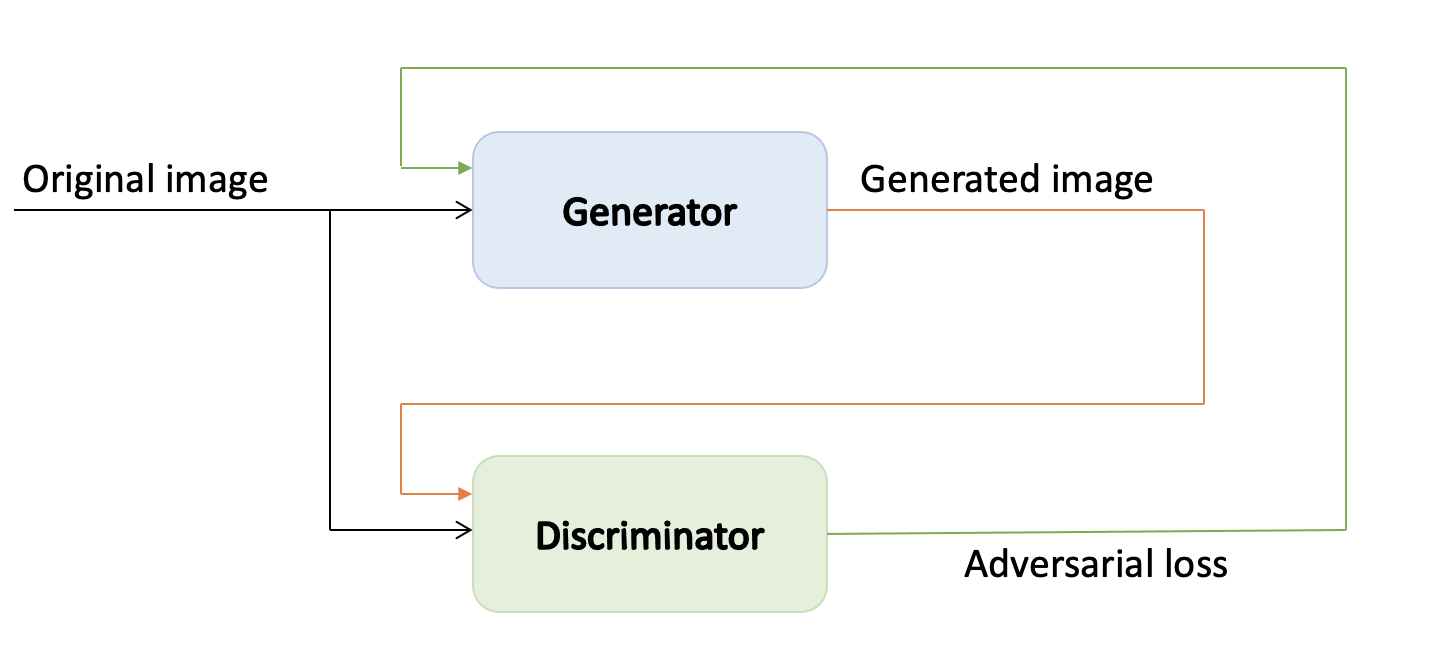}}
\caption{GAN framework proposed by Goodfellow et al. \cite{goodfellow2020generative}.}
\label{fig:GAN}
\end{figure}

\hfill\\
Generative adversarial attacks employ generative models, such as a GAN (Generative Adversarial Network), to create adversarial examples.  GAN attack can indeed be classified as an adaptive adversarial attack when the discriminator component of the GAN is set to the target model. 
\citeauthor{goodfellow2020generative} \cite{goodfellow2020generative} leverages two neural networks: a generator (G) and a discriminator (D), as shown in Fig.\ref{fig:GAN}. The generator (G) is trained to generate synthetic images that are similar to the target images, but that are misclassified by the target model, while the discriminator (D) is trained to distinguish between real images and those synthesized by the generator. The adversarial loss is used to measure how successful the generator is in deceiving the discriminator in a GAN.  Let us suppose that  $I$ is the generator's input and that $G(I)$ is the corresponding generated image. The adversarial loss is computed as $log(1 - D(G(I)))$, where $D(G(I))$ represents the probability that the generated image $G(I)$ originates from real data. Minimizing the adversarial loss allows the generator to learn to produce visually realistic images that deceive the discriminator.\\
In an adaptive GAN attack, the attacker sets the discriminator to mimic the target model they intend to attack. The attacker then trains the generator to generate adversarial examples that can fool the discriminator, which is essentially the target model itself. During the attack, the attacker iteratively adjusts the parameters of the generator based on the discriminator's feedback, allowing them to adapt their attack strategy in response to the target model's defenses. By continually training the generator and receiving feedback from the discriminator, the attacker can refine the adversarial examples to maximize their effectiveness and increase the chances of successful evasion.\\
The authors show that their GAN is highly effective at fooling a range of image classifiers. Moreover, they demonstrate that their generative adversarial attack is effective even in a black-box setting, where the attacker has limited access to the target model.
\subsection{Existing defenses}
One approach to defending against adversarial examples is to improve the robustness of the classifier by enhancing its training process \cite{shaham2015understanding}. One intuitive strategy to achieve this is by incorporating adversarial information during training, commonly referred to as adversarial training.
\citeauthor{madry2017towards} \cite{madry2017towards} proposed the use of Projected Gradient Descent (PGD) for adversarial training in deep learning. The authors demonstrated that adversarial examples generated by PGD are more representative of adversarial examples generated with first-order methods. The authors showed that the deep learning model shows good robustness on MNIST and CIFAR10 datasets. Nonetheless, the training process was computationally expensive, making it challenging to apply this approach to large datasets. Adversarial training can be effective against adversarial attacks. However, there is no guarantee about the models' safety since it depends on the attack used. It is important to note that our proposed defense mechanism does not fall under the category of adversarial training and does not include adversarial examples in training.

Further defense strategy involves detecting adversarial examples through the use of dedicated classification networks and statistical-based detectors. Researchers have proposed add-on neural network  detectors to enhance the robustness of neural network classifiers against adversarial attacks. These detectors are often designed to detect the presence of adversarial perturbations in the input prior to classification by the network. For instance, \citeauthor{metzen2017detecting} \cite{metzen2017detecting}, \citeauthor{gong2017adversarial}  \cite{gong2017adversarial}, and \citeauthor{lu2017safetynet}, \cite{lu2017safetynet}  proposed the use of deep neural network (DNN) binary detectors to distinguish between legitimate and adversarial inputs. These detectors are  trained using both normal and adversarial examples. However, it has been observed that these detection approaches may not generalize well across various attack parameters and generation processes, highlighting the need for more robust defense mechanisms. Other researchers have proposed statistical-based detection techniques for identifying adversarial examples. For instance, \citeauthor{feinman2017detecting} \cite{feinman2017detecting} utilized Bayesian uncertainty analysis to detect adversarial examples. The authors claimed that the uncertainty associated with adversarial examples is greater than that of legitimate data. To this end, they proposed a Bayesian neural network for estimating the uncertainty of input data and subsequently distinguishing between adversarial and legitimate data based on uncertainty estimation.
Unlike the aforementioned approaches, our proposed approach is orthogonal to this branch of work. Our proposed method does not require adding any additional defense framework to mitigate adversarial attacks.

Defensive distillation \cite{papernot2016distillation}, trains the classifier in a manner that renders it extremely challenging for gradient-based attacks to generate adversarial examples directly on the network. Specifically, defensive distillation employs distillation training techniques \cite{hinton2015distilling} to mask the gradient between the pre-softmax layer (logits) and softmax outputs, thereby enhancing the robustness of the classifier against adversarial attacks. Nevertheless, \citeauthor{carlini2017towards} \cite{carlini2017towards} has demonstrated that it is relatively straightforward to circumvent defensive distillation using one of three strategies: (1) selecting a more suitable loss function, (2) computing the gradient directly from the pre-softmax layer instead of from the post-softmax layer, or (3) attacking a network that is easier to compromise first and then transferring the attack to the distilled network. In the context of white-box adversarial attacks, when the attacker has knowledge of the parameters of the defense network, it becomes challenging to prevent the generation of adversarial examples that can successfully defeat the defense mechanism. Thus, in this study, we propose an alternative approach that focuses on the gray-box model. Our approach involves introducing a randomization technique to make it more difficult for the attacker to generate effective adversarial examples.

A different family of adversarial defenses that have merged lately is  input transformations. Input transformation aims to increase the robustness of the models against adversarial attacks independently of the attacks.  \citeauthor{xie2017mitigating}\cite{xie2017mitigating} suggested utilizing two randomization techniques to mitigate the effects of adversarial perturbations: random resizing and random padding of input images. Random resizing involves resizing the input image to a randomly selected size, while random padding entails adding zero pads around the image in a random fashion. While these transformations aim to reduce the impact of adversarial perturbations, they may not be effective against powerful attacks.  It should be noted that our defense approach differs from  \cite{xie2017mitigating} method in that we do not employ random resizing or padding for input randomization.\\
\citeauthor{raff2019barrage}\cite{raff2019barrage} utilized a technique where a varying number of transformations (such as noise injection, FFT perturbation, JPEG Noise, etc.) were applied to images in a random sequence during both training and testing stages.  This method showed promising results in defending against adversarial attacks, but it also resulted in a considerable decrease in the deep learning classifier's accuracy. It is worth noting that our proposed approach employs a distinct method of input randomization. Furthermore, our proposed randomization method has a minor impact on the accuracy of the classifier.\\
\citeauthor{liu2018towards} \cite{liu2018towards} introduced a defense mechanism called random self-ensemble, which utilizes random noise. They incorporated a noise layer before each convolutional layer of the deep learning network during training and testing. However, they later acknowledged in another work that blindly adding noise to each layer is not an effective way to incorporate noise \cite{liu2018adv}. Similarly,\citeauthor{you2019adversarial} \cite{you2019adversarial}  also added a noise layer to the convolutional network. It is worth noting that these prior works introduced a random noise layer to the classifier, which is distinct from our proposed defense method. Our approach does not involve modifying the classifier's architecture.\\
\citeauthor{taran2019defending} \cite{taran2019defending}  proposed a defense mechanism utilizing key-based randomization. They introduced a multi-channel architecture that aggregates classifiers' scores. Each channel introduces randomization to the input data based on a random key $k$. Multiple classifiers are trained using the randomized inputs, and the final decision is based on the aggregated outputs of the classifiers. The secret key k is shared between the training and testing phases and is unknown to the attacker. In contrast, our proposed defense technique trains classifiers to classify inputs in the projected domain and does not revert the input to the original domain. Additionally, our approach does not involve aggregating outputs from different trained classifiers.

\section{Threat Model}
\label{sec:threat}
In this section, we outline the threat model used in this work  for assessing the security of the proposed defense mechanism against potential attacks. \\
\textbf{\em{Threat Model}} We assume a gray-box threat model where the adversary has  limited knowledge about the classifier it aims to attack, referred to as the target model. This is a common assumption in previous work \cite{meng2017magnet,guo2017countering}.
The adversary has direct access to the target model architecture and parameters.  Moreover, the adversary has access to the training data used to train the target model. However,  the adversary is not aware of the defense technique deployed in the target model and does not have access to the random  images used by the defense model. \\\\
\textbf{\em{Adversary's Goal}} The goal of the attacker is to generate adversarial examples that mislead the target model. In adversarial attacks, the attacker will either aim to reduce the classification accuracy of the target model (untargeted attacks) or target a specific class (targeted attacks). In this work, we assume that the adversary aims to perform targeted attacks on a target model. To do so, the adversary will create adversarial examples that can be either a modified version of  the input or a completely new input designed to exploit vulnerabilities of the target model, with the goal of having its generated input classified as a specific class of the attacker's choice, rather than just any arbitrary class. For instance, in the context of the MNIST dataset, which consists of images of handwritten digits, the attacker aims to misclassify an image of the digit "3'' as the digit "8'' rather than any arbitrary class (e.g., 0, 1, etc.).
\\\\
\textbf{\em{Adversary's capabilities}}
While the adversary does not have access to the random images used by the proposed 
defense, it has access to the model's output with respect to the random images. The adversary can use the output of the target model to observe the model and generate adversarial examples.

\section{Our solution}
\label{sec:dtl_sol}
In the following subsections, we first report our solution at a glance, and later  we provide a detailed description of the proposed defense mechanism. 
\subsection{Solution at a glance}
The proposed defense mechanism is designed to enhance the robustness of deep learning models against adversarial attacks by leveraging the power of randomness. 
 The core idea behind the defense mechanism is to use random hyperspace projection to project the input dataset to another dataset (hyperspace).
To this end, the proposed mechanism uses random images, to project the input training data into another hyperspace. Thus, we create a number of different projected datasets. The projected training datasets are then used to train the target model. This training process yields to several trained classifiers. The resulting trained classifiers have different decision boundaries and have learned different features. 
During the testing phase, the defense mechanism applies a random projection at every test step. It first generates/retrieves a random image used in the training phase and projects the input using the corresponding random image. Then it retrieves the trained classifier with respect to the random image chosen and predicts the final output using the transformed input and the selected classifier---the one trained on that projection. This approach enhances the robustness of the classifier to attacks since  the attacker is not aware of the random images used by the target model. 
Hence, it is not aware of which trained classifier we have used for predicting the output of its adversarial example.  Furthermore, the diversity of classifiers makes it more challenging for an attacker to generate successful adversarial examples, as the attacker would need to craft an attack that works against all or most of the classifiers. Even if the attacker has crafted a successful adversarial example that was able to mislead one of the trained classifiers, the adversarial example will succeed on the rest of the set of trained classifiers with low probability.

\subsection{Detailed Solution}
In this work, we aim  to design a defense mechanism that is able to protect the DNN classifier from adversarial attacks while maintaining a high classification accuracy.
Our solution is based on hyperspace projection and shuffling. \\

{\em \textbf{Hyperspace projection }}Hyperspace projection is a technique that maps an input data into a new data point. In this context, we project the input image into a new image by  applying the XOR operation to the pixel values of the input image and a random image.

Formally, let $X$ be a dataset of $|X|=x$ data points, that is: $X=\{X_0,X_1,\ldots,X_{x-1}\}$ and let $R$ be a set of $|R|=r$ random generated images, $R=\{R_0,R_1,\ldots,R_{r-1}\}$. Let $X_i$ be an $(n \times m)$ bit matrix, with  $X_i(l,p)$ identifying the bit at the $l^{th}$ row ($ 0\le l \le n-1$) and $p^{th}$ column ($ 0\le p \le m-1$). Let $R _j$ be a randomly generated $(n \times m)$ bits image ( $R_j\xleftarrow{\$} \{0,1\}^{n \times m}$), with an element of  such a matrix being  represented, as per $X_i$, by $R(l,p)$.
The hyperspace projection: $$p(\circ,\circ):\{0,1\}^{n \times m} \rightarrow \{0,1\}^{n \times m}$$ taking in input $X_i$ and $R_j$ produces a new image $X'_{i,j}$ ($X'_{i,j} \leftarrow p(X_i,R_j)$) that is  defined as:
\begin{equation}
    X'_{i,j}(l,p)=X_i(l,p) \oplus R_j(l,p)
    \label{eq:xor}
\end{equation}
where $\oplus$ is the exclusive-OR binary operator. 
Hence, the element-wise XOR operation between the input image  matrix $X_i$ and the random image matrix  $R_j$ defines a hyperspace projection that maps each element of $X_i$ onto a new element $X'_{j,i}$.
It is worth noting that the projection, if considered on any single image in isolation, maps that image to a random element of the space. Though, a key observation is that if the image $R_j$ is the same for a set of images, the projection will preserve certain properties of the original images.
For instance, consider a set  $\mathcal{S}_i$ of $\mathcal{L}$  bits in $X_i$ that represent, for the classifier, a relevant  feature of the image. The projection will  transform that set in $\mathcal{S}'_i$. 
However, for any other image other than $X_i$ (say $X_j$), if that image sports the set $\mathcal{S}_j$ with $\mathcal{S}_j=\mathcal{S}_i$, that very set $\mathcal{S}_j$ will be transformed into $\mathcal{S}'_{j}$, with $\mathcal{S}'_{j} = \mathcal{S}'_{i}$, since 
$\mathcal{s}'_j(l,p)=\mathcal{s}'_i(l,p)= \mathcal{s}_i(l,p) \oplus R_j(l,p)=\mathcal{s}_j(l,p) \oplus R_j(l,p)$.
Hence, preserving in the hyperspace the similarities among the images, for that given feature, that existed in the departure domain.\\
In our solution, we project all the elements of the dataset into a hyperspace using a random image $R$. We  repeat the process $r$ times, using each time a different random image $R_i$, $i=0,\ldots, \mbox{r-1}$.
We claim, on the one hand, that the projection does not alter the ML capabilities to extract and leverage the necessary features that support the classification task and, on the other hand,  that the suggested hyperspace projection can be effective as a defense mechanism against adversarial attacks since it transforms the input data into a new space that is different from the original space.  These two claims are proved in the sequel of the paper\footnote{What just stated holds true under the assumption that an $R_j$ is taken uniformly at random from the space $\{0,1\}^{\{n,m\}}$; in our case, we resort to a PRNG generator to produce $R_j$, hence the hypothesis is reduced to the quality of the PRNG.
}.\\
The dataset hyperspace projection: $$dp(\circ,\circ):\{0,1\}^{x\times n \times m} \rightarrow \{0,1\}^{x\times n \times m}$$ takes in input a  dataset $X=\{X_0,X_1,\ldots,X_{x-1}\}$ 
and an element   $\in~\{0,1\}^{n\times m}$---such as a random image $R_j$---and produces a new projected dataset $X' \leftarrow dp(X,R_j)$, 
that is defined as:
\begin{equation}
    dp(X,R_j)=\{p(X_0,R_0), \ldots,~p(X_{x-1},R_{x-1}) \} 
    \label{eq:dp}
\end{equation}

The intuition behind why our solution could be successful against an adversary (as it will be shown later on), is that when an adversary attempts to create an adversarial example, he is essentially trying to find a perturbation to the original input data that causes a misclassification by the deep learning model. 

In the case of non-iterative adversarial attacks, the adversary has one chance to generate the perturbation to craft a successful adversarial example. The adversary is not aware of the hyperspace projection applied by the defense mechanism. Hence, he will try to find the perturbation that deceives the model in the original space.  However, when the adversarial example is projected to a different hyperspace using the random image, the perturbation that may be  effective in the original space are unlikely to be effective anymore in the projected space. Specifically, even if the adversary manages to find a successful adversarial example in the original input space, it is unlikely to cause a misclassification in the projected space as it leads to different decision boundaries, which may not be crossed by the  projected adversarial perturbation.\\
However, in iterative attacks, the adversary is able to query the model multiple times and make changes to the input based on the feedback received from the model. In each iteration, the adversary's perturbation is updated based on the current model's feedback. However, the hyperspace projection is also updated at each iteration by using a different random image. Thus, the model is changed accordingly at every iteration. This means that the feedback received by the adversary after each query will be irrelevant for the next iteration as the model's boundaries have been changed. Therefore, it becomes difficult for the adversary to generate a perturbation that is effective across multiple iterations, decreasing the effectiveness of the adversarial attack in the short run.

\begin{figure}[htbp]
\centerline{\includegraphics[width=\linewidth]{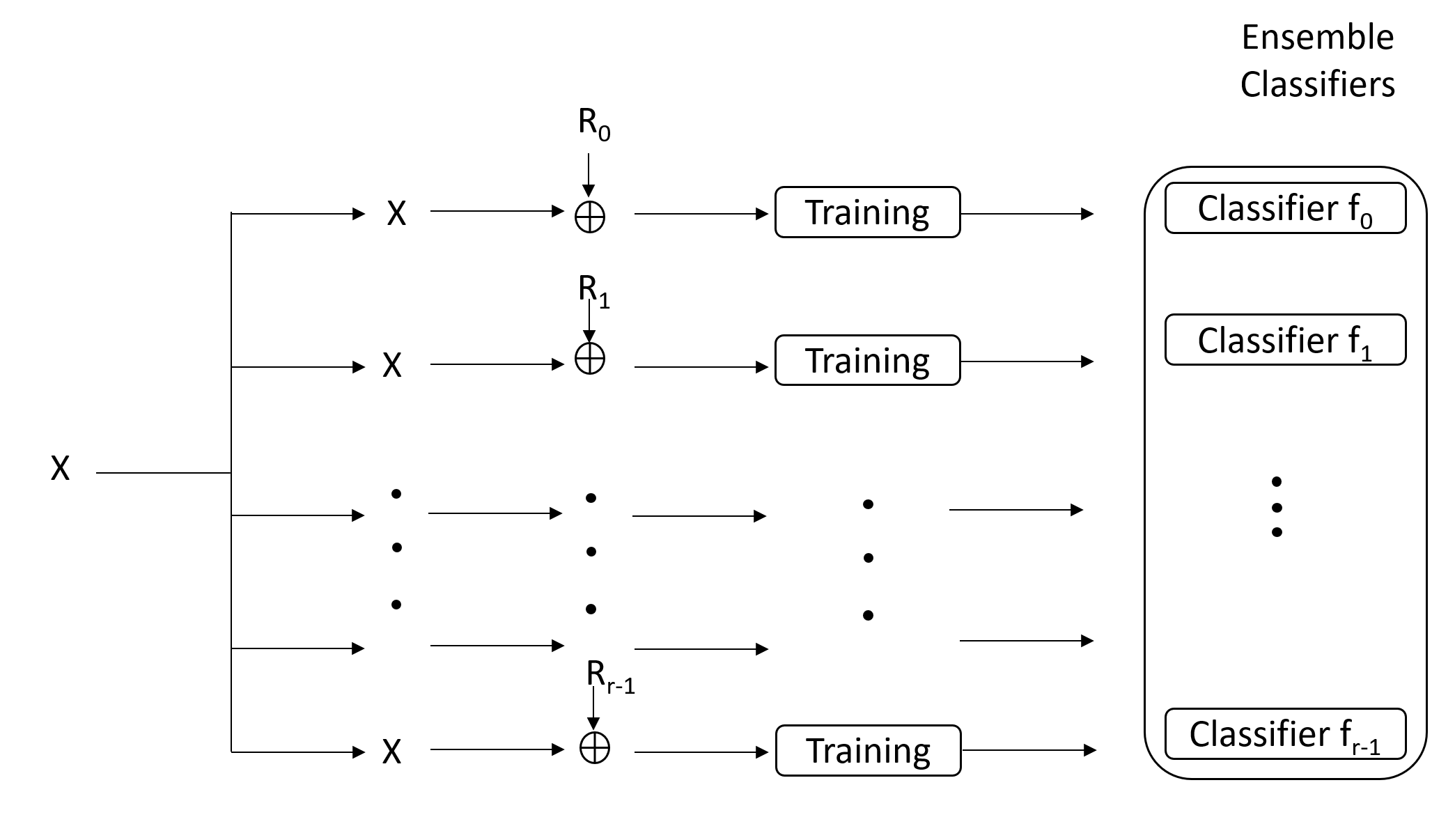}}
\caption{Proposed defense mechanism in the training phase.}
\label{fig:training}
\end{figure}

{\em \textbf{Proposed Solution}} The proposed defense method comprises two phases: The training phase and the testing phase.\\
The main objective of the training phase is to create a train a large number of classifiers using the random projected datasets as shown in Fig. \ref{fig:training}. In order to achieve this objective, the proposed defense mechanism follows several steps. First, we create a set of random images $R=\{R_0, R_1, R_2.\ldots, R_{r-1}\}$.
These  images should be kept confidential with respect to the attacker. They are created by generating random pixels of the training image size. This can be accomplished by assigning a random grayscale value to each pixel within the image. 

This approach produces completely random images, which can be employed in the projection of the input dataset.  Moreover, the resulting images, obtained from this approach, are diverse and unpredictable, further enhancing the defense mechanism's robustness against potential adversarial attacks. Secondly, in order to generate the projected datasets $X'_j$,  each pixel of the random images, $R_j, j=0,\ldots,r-1$, is XORed, bit by bit, with the corresponding value in the input training data $X$, as shown in Equation. \ref{eq:dp}. In this work, we use the logical operator XOR to project the dataset as it can help create a non-linear mapping that preserves the features of the original data. Moreover, the use of XOR in this context can be seen as a form of data encryption or obfuscation, as the projected image is difficult to reverse-engineer without knowledge of the original  random image---other than being an  efficient operation. \\
Finally,  we train a DNN classifier $f$  using the $r$ projected datasets. We obtain a set of trained classifiers $(f_0, f_1, \dots, f_{r-1})$  that we will refer to as ensemble classifiers. Although we have used the same classifier $f$ and the same hyperparameters, the trained ensemble classifiers are not identical since they are trained on the different projected datasets $X'_j, j=0,\ldots,{r-1}$. Each classifier $f_j,j=0,\ldots,{r-1} $ will learn different patterns in the data based on the unique characteristics of its training set. Therefore, even though the classifiers share the same model structure and tuning parameters, their learned weights and biases will differ, resulting in unique predictions for a given input.  The pseudocode for the training phase is depicted in Alg.\ref{alg:training}.

\begin{algorithm}
\caption{Training phase( )}
\label{alg:training}
\begin{algorithmic}[1]
\Require a training dataset $X$
\Ensure a projected dataset $X'_j$
\State Define a classifier $f$.
\State Define the original training dataset $X$.
\State Define a set $R$ of different random images that will be used for projection.
\For {each random image $R_j$ in the set $R$}
\State Create a new training dataset $X'_j$ by XORing the original 
\NoNumber{training dataset $X$ with the XORing random  image }
\NoNumber{$R_j$.}
\State Train a classifier $f$ using the new training dataset $X'_j$.
\State Save the trained classifier $f_j$.
  \EndFor

\end{algorithmic}
\end{algorithm}

\begin{figure}[htbp]
\centerline{\includegraphics[width=\linewidth]{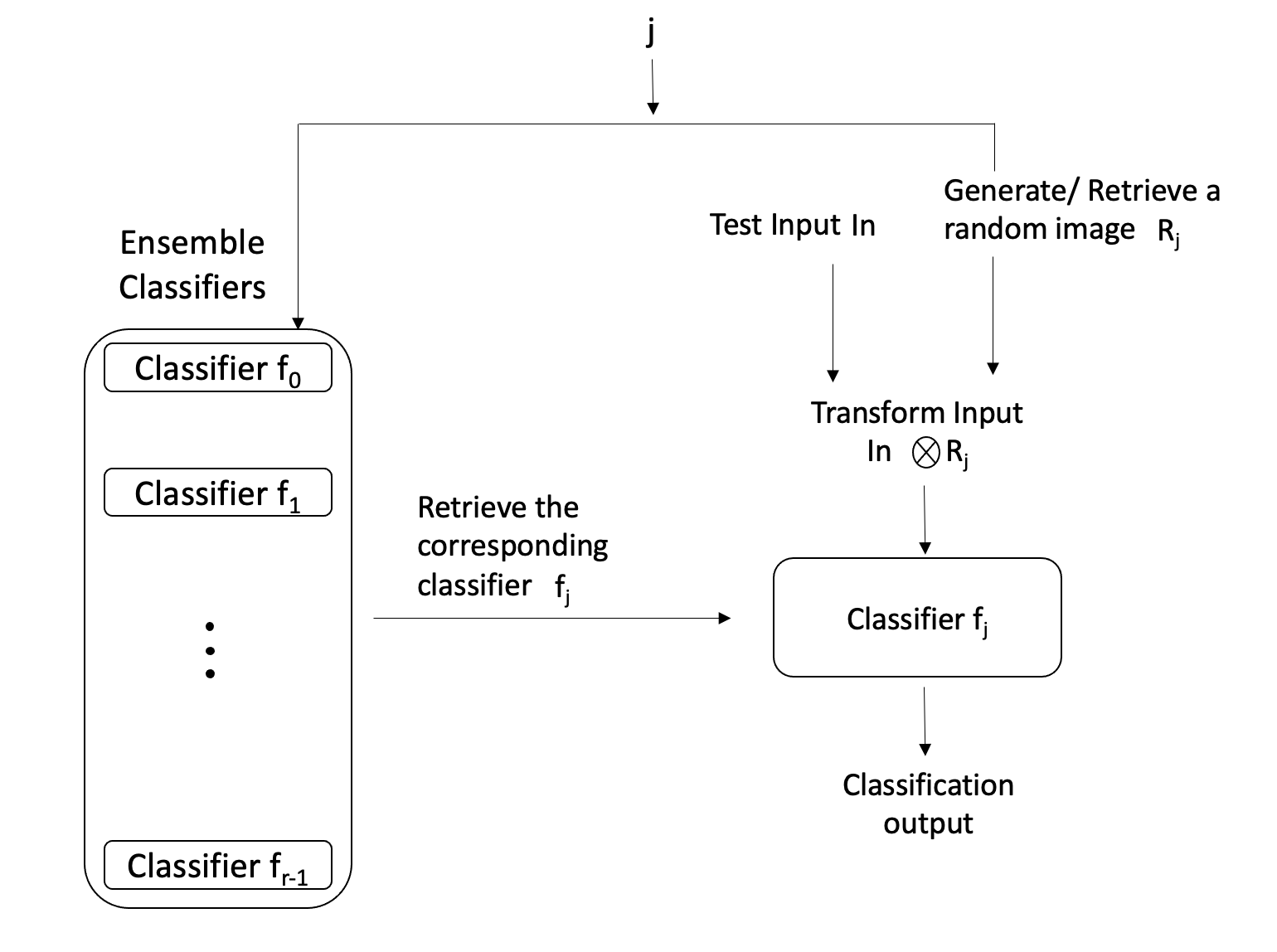}}
\caption{Proposed defense workflow in the test phase. }
\label{fig:testing}
\end{figure}

The workflow of the proposed defense mechanism is depicted in Fig. \ref{fig:testing}. During every testing iteration,  we randomly generate/ retrieve a random image $R_j$ from the available set of random images, $R$. We apply the input transformation using the chosen random image $R_j$. This will project the test input from its original hyperspace to the chosen hyperspace. Secondly, we retrieve the corresponding trained classifier $f_j$ from the ensemble of classifiers. This ensures that  the model is evaluated on inputs that have been transformed in the same way as the training data. Finally, the classifier will output its prediction. As mentioned in Section. \ref{sec:threat}, the attacker will have access to the prediction output, but it does not have access to the output of the defense method, i.e., the random projection. Moreover, the defense method uses a different random image $R_j$ at every iteration, effectively changing the hyperspace and the classifier used each time. The pseudocode of the proposed testing strategy is depicted in Alg. \ref{alg:testing}.\\
The proposed defense strategy prevents the attacker that launches iterative attacks from learning the projection by observing the model's output. Furthermore, it significantly increases the difficulty of generating successful attacks, as the output at every iteration will correspond to a different random image transformation. Hence, it is difficult for the attacker to learn all the hyperspace projections and craft successful attacks.  This approach further increases the randomness and unpredictability of the defense mechanism, making it harder for the attacker to find a successful attack strategy.  However, the number of random images used depends on the defender's choice and the computational resources available. In the following sections, we will study the impact of the number of random images used on the proposed defense's robustness against adversarial attacks.

\begin{algorithm}
\caption{Testing phase( )}
\label{alg:testing}
\begin{algorithmic}[1]
\Require an input image $In$
\Ensure classification label $l$ 
\For {each input $In$}
\State Generate a random index $j\in{0,\ldots,r-1}$

\State Generate/Retrieve the random image $R_j$

\State Project the input $In$ to the selected hyperspace using 
\NoNumber{the random image $R_j$.}
\State Retrieve the trained classifier $f_j$ corresponding 
\NoNumber{to the selected random image.}
\State Use the trained classifier $f_j$ to compute  a prediction ($l$).
\State return($l$).
\EndFor

\end{algorithmic}
\end{algorithm}

\begin{figure*}[htbp]
\centerline{\includegraphics[width=0.8\linewidth]{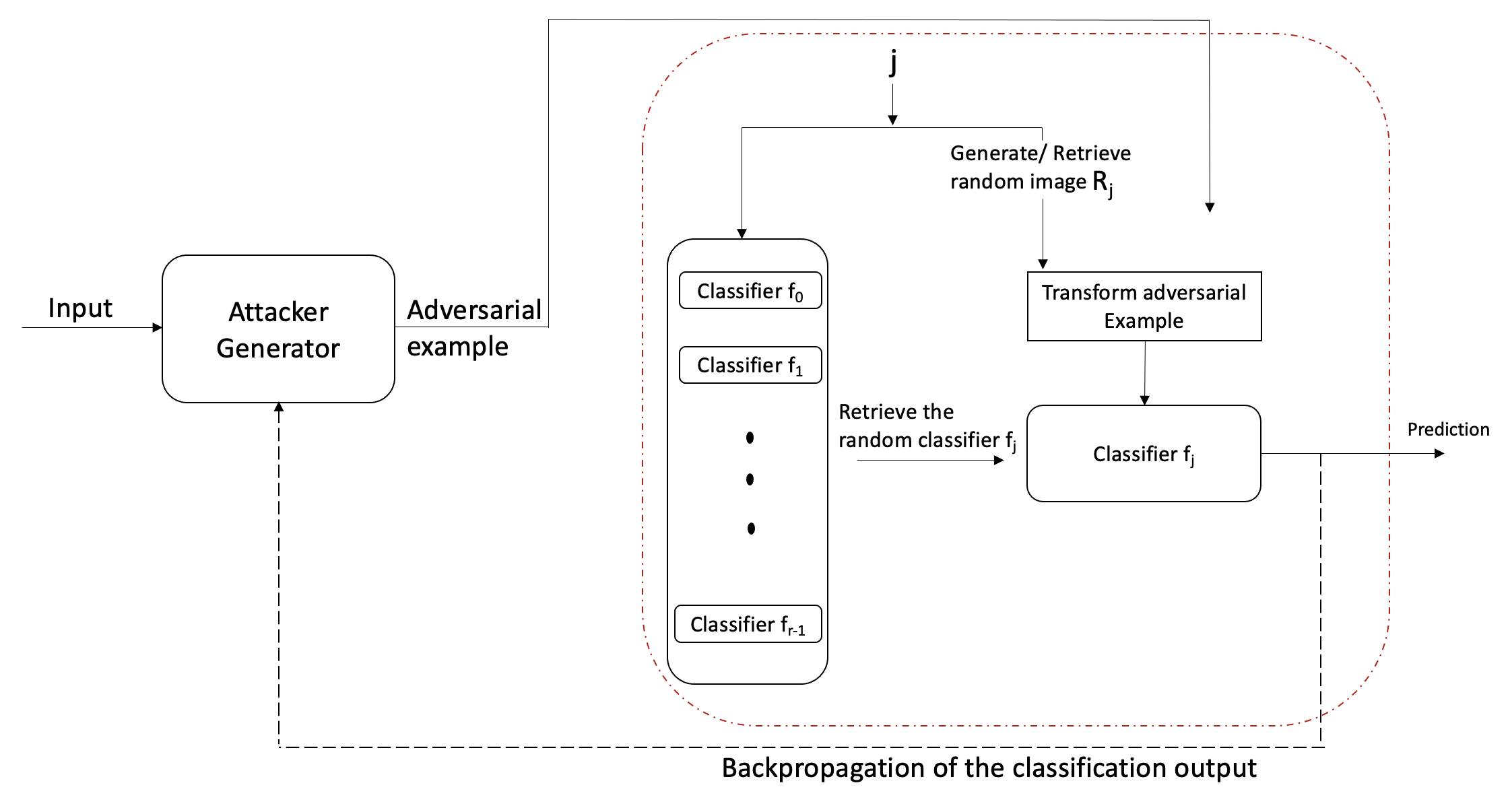}}
\caption{Attack overview}
 \label{fig:attack}
\end{figure*}

After describing the defense model, we now turn our attention to how it works in the presence of adversarial attacks. The attacker aims to generate adversarial examples that cause the target model to make a specific incorrect prediction. To achieve this, The attacker may use various techniques to generate its adversarial example. We consider a game between the defender and the attacker, as shown in Fig. \ref{fig:attack}. The attacker can either add a small perturbation to the original input or generate a new one.  From the defender's perspective, the classifier  has been trained with the hyperspace projection defense mechanism, but the attacker is not aware of this fact.  Therefore, In the case of non-iterative attacks, the attacker cannot predict the amount of noise added to the input data during inference. Thus,  the generated adversarial examples  will likely be ineffective against the model since they do not consider the random noise added to the input data during inference. In the case of iterative attacks,  the attacker tries to guess the random image by observing the model output with respect to its input. However, in each iteration of the attack, the defense mechanism involves a new random image resulting in a new projection of the input data, making it much more difficult for the attacker to try to estimate the random image and generate effective adversarial examples.

\section{Implementation and Evaluation}
\label{sec:exp}
In this section, we evaluate the effectiveness of the proposed defense mechanism and compare it against competing solutions. 
To this end,  we conduct several experiments adopting as a reference dataset  the MNIST  one. 
\subsection{Setup}
We evaluated our proposed defense mechanism using two datasets: MNIST \cite{yann1998mnist}. We selected 55000 examples for the training set, 10000 examples for the testing set, and 5000 examples for the validation set, \cite{meng2017magnet}. The architecture of the classifier to be protected is described in Table \ref{tab:architecture}. We trained the model using the settings described in Table \ref{tab:training_par}, and we achieved an accuracy of $99.1\%$.\\
We have created $r=64$ random images as described in Section \ref{sec:dtl_sol}. Since the MNIST dataset consists of $28 \times 28$ grayscale images, the random images will also be grayscale and have a size of  $28 \times 28$ pixels. The random images, $R_j, j=0,\ldots, 63$, are created by generating random pixels. The random pixels are generated one by one using a random number generator to generate a random value between 0 and 255 for each pixel, with 0 being black and 255 being white. The generated random  images are securely stored and are assumed to be kept private with respect to the adversary. The random images are then used to project the MNIST dataset into different hyperspaces. We train the classifier described in  Table \ref{tab:architecture} using the $r$ projected dataset adopting the same settings described in  Table \ref{tab:training_par}. The trained models are securely stored as well to avoid the need to retrain the model during the testing phase.  This approach ensures a fast and smooth procedure and avoids the time-consuming training process.
\begin{table}[htbp]
\centering
\caption{Architecture of the target classifier}
\label{tab:architecture}
\begin{tabular}{ll}
\hline
\multicolumn{2}{c}{MNIST Classifier} \\ \hline
Layer             & Parameters       \\ \hline
2DConv.ReLU       & 3x3x32           \\ \hline
2DConv.ReLU       & 3x3x32           \\ \hline
MaxPooling        & 2x2              \\ \hline
2DConv.ReLU       & 3x3x64           \\ \hline
2DConv.ReLU       & 3x3x64           \\ \hline
MaxPooling        & 2x2              \\ \hline
Flatten           & -                \\ \hline
Dense.ReLU        & 200              \\ \hline
Dense.ReLU        & 200              \\ \hline
Softmax           & 10               \\ \hline
\end{tabular}
\end{table}
\begin{table}[htbp]
\centering
\caption{Training parameters of the target classifier}
\label{tab:training_par}
\begin{tabular}{ll}
\hline
Parameter           & Value \\ \hline
Optimization method & SGD   \\ \hline
Learning Rate       & 0.01  \\ \hline
Batch size          & 128   \\ \hline
Epochs              & 10    \\ \hline
Loss function       & categorical crossentropy    \\ \hline
\end{tabular}
\end{table}

For the rest of the section, we will evaluate the robustness of the proposed defense mechanism under gray-box attacks. We first study the impact of the defense mechanism on the model's accuracy. We then evaluate the robustness of the defense model under optimization-based attacks and generative attacks. Finally, we conduct a study on the impact of the number of random images used on the robustness of the proposed defense.\\
The adversarial attacks can be divided into two categories: (1) untargeted attacks; and, (2) targeted attacks. In untargeted attacks, the attacker  aims to create adversarial examples that misclassify the  target model, without any specific target or goal in mind. The attacker's objective is simply to cause a misclassification. On the contrary, in targeted attacks, the attacker chooses a specific class and creates adversarial examples that  cause the target model to misclassify them  into this particular class. As expected,  targeted attacks are more difficult to defend against compared to untargeted attacks. Hence, robustness to targeted attacks provides a more rigorous evaluation of the defense mechanism. Moreover, targeted attacks can reveal any weaknesses or vulnerabilities in the defense mechanism that attackers can exploit, enabling us to refine and improve the defense mechanism. Therefore, testing with targeted attacks can provide a more comprehensive and accurate assessment of the defense mechanism's effectiveness in real-world scenarios.
\subsection{Impact of the  proposed defense mechanism on classification accuracy}
One of the objectives of  our defense technique is to preserve good classification accuracy even when classification is performed on a  projected hyperspace.
To this end, in order to evaluate whether the presence of the proposed defense technique affects the model's classification accuracy, we trained the target model described in  Table \ref{tab:architecture}  using $r$ randomly generated images, with $r=64$. 
The classification results of the $r=64$ trained model are reported in Table \ref{tab:acc}. We can observe that the model's accuracy has slightly dropped by $1.01\%$, on average.
Despite the slight decrease in classification accuracy, the model deploying our proposed defense method is still able to maintain a high level of accuracy under the proposed transformation. Indeed, the classifier trained on the projected datasets was able to achieve an accuracy of $97.92\%$ on average. The highest accuracy that the classifier achieved is $98.09\%$.  This experimental result demonstrates that our defense technique is highly effective at maintaining classification accuracy, unlike the state-of-the-art input transformation defense technique \cite{raff2019barrage}. It is worth noticing that it would also be possible to further reduce the accuracy drop by further  optimizing the proposed defense training parameters. However, given the already more than tolerable drop in accuracy, to keep focus, and for the sake of space, we leave this task for future work.

\begin{table}[htbp]
\centering
\caption{Classification accuracy of the original classification  model and the proposed model}
\label{tab:acc}
\begin{tabular}{c|cccc|}
\cline{2-5}
\multicolumn{1}{l|}{}                                    & \multicolumn{4}{c|}{Validation accuracy}                                                             \\ \hline
\multicolumn{1}{|c|}{Model without  defense} & \multicolumn{4}{c|}{99.1\%}                                                                          \\ \hline
\multicolumn{1}{|c|}{\multirow{2}{*}{Proposed model}}    & \multicolumn{1}{c|}{Min}     & \multicolumn{1}{c|}{Max}     & \multicolumn{1}{c|}{Avg}     & Std     \\ \cline{2-5} 
\multicolumn{1}{|c|}{}                                   & \multicolumn{1}{l|}{96.97\%} & \multicolumn{1}{c|}{98.09\%} & \multicolumn{1}{c|}{97.92\%} & 0.019\% \\ \hline
\end{tabular}

\end{table}

\subsection{Proposed defense mechanism performance under optimization-based attacks}
\textbf{\em{Attacks configuration}} We evaluate the proposed defense mechanism on four representative optimization-based adversarial attacks: FGSM, IGSM, PGD, and C\&W. We have used these attacks to generate adversarial attacks against the proposed defense mechanism on MNIST dataset. The attack configurations are reported in Table \ref{tab:conf-opt}. \\

\begin{table}[htbp]
\centering
\caption{Configuration of optimization-based adversarial attacks}
\label{tab:conf-opt}
\begin{tabular}{cc}
\hline
{\em Attack method} & {\em Configuration} \\ \hline
FGSM          &     $\epsilon$=0.3          \\ \hline
IGSM          &        $\epsilon$=0.01, number of iterations=100       \\ \hline
PGD           &           $\epsilon_{max}$= 0.3, number of iteration=100, $\epsilon=0.01$    \\ \hline
C\&W          &         \begin{tabular}[c]{@{}c@{}}Binary step size=9, max iterations=1000,  \\learning rate=0.01, abort early=True\end{tabular}  \\ \hline
\end{tabular}
\end{table}

We run each of the aforementioned adversarial attacks on the MNIST classifier twice, one without defense and one with the proposed defense mechanism. In this experiment, we evaluate the proposed mechanism using just two random images (that is, $r=2$).
More specifically, we generate a small set of random images $R$, with $r$ representing the number of random images contained in the subset, such that $ |R| = r\}$.  At each iteration, the defense mechanism uses different random images from the subset $R$. In the following, we will use a subset with two random images $r=2$. The impact of the number of random images used by the defense method will be studied later.  The iterative attacks, i.e., IGSM, PGD, and C\&W, interact with the target model through multiple iterations. At each iteration, the attacker adjusts the attack parameters and perturbations based on the feedback received from the classifier $r_i$ used by the defense mechanism to find the optimal direction that increases the model's loss and maximizes the chances of misclassification. Hence, in this scenario, the iterative attacks  exhibit adaptive characteristics by adapting their attack parameters or perturbations accordingly at each iteration to exploit the specific defense mechanism employed.\\
Table \ref{tab:ASR-opt} shows the average success rate of the adversarial attacks with and without  the proposed defense mechanism in place. 
In this experiment, as we evaluate the defense mechanism on targeted attacks, we iteratively test it on every label and report the attack success rate across all labels. This process is repeated five times, and the reported success rate is an average of these experiments.  The results show that without the defense mechanism, all four attack methods are very effective, with success rates ranging from $56.87\%$ to $92.1\%$. This means that the model is vulnerable to adversarial attacks. However, when using the proposed defense mechanism, we notice that the attack success rate has significantly reduced. Indeed, the adversarial attacks' success rate is below $10\%$ for all the optimization-based adversarial attacks, with $FGSM$ being the lowest with an attack success rate of $1.64\%$ and $C\&W$ being the highest with an attack success rate of $9.45\%$. These results suggest that the defense mechanism is highly effective at reducing the success rate of attacks. Specifically, for all four attack methods (FGSM, IGSM, PGD, and C\&W), there is a significant improvement in accuracy when the defense mechanism is applied. The improvement ranges from $89.7\%$ to $97.1\%$, indicating that the defense mechanism is highly effective at reducing the impact of adversarial attacks. \\
FGSM attack has achieved the lowest attack success rate because of the nature and the design  of the attack. FGSM is a simple one-shot attack that generates adversarial examples by adding a small perturbation to the input data in the direction of the gradient of the loss function with respect to the input. However, the proposed defense mechanism adds pseudo-random noise to the input data during inference, which makes it more difficult for an attacker to generate effective adversarial examples. When random noise is added to the input data, the gradient direction of the loss function with respect to the input becomes less predictable, which makes it more difficult for the attacker to generate perturbations that will result in successful adversarial examples. In contrast to FGSM, iterative attacks such as IGSM, PGD, and C\&W use a series of iterations to optimize the perturbations that will be used to generate adversarial examples. However, even with iterative attacks, the attack success rate is still low ($<7\%$). These findings confirm that the proposed defense mechanism disrupts the optimization process of adversarial attacks, making it more difficult for the attacker to generate successful examples.\\
We would also like to highlight  that in the context of targeted attacks on the MNIST model with 10 labels, the model is trained to output one label out of the 10 labels ($0,\dots,9$). This means the model will always produce a prediction, even if the input is not a valid MNIST or it is a random image.
In this experiment,  the attacker is attempting to create an example that the model will misclassify as a specific target label; however, because the MNIST model is designed to predict one of 10 labels, we notice that an invalid or truly random input can still be classified as one of these labels with $10\%$ probability, due to the design of the MNIST model. In a sense, as it can be seen from the results reported in Table \ref{tab:ASR-opt}, our solution degrades the attacker's capability well below what one would obtain with a plain random choice. \\
To summarize,  we have shown above that
 the proposed defense mechanism can be effective against both non-iterative (FGSM) and iterative (PGD, IGSM, and C\&W) optimization-based adversarial attacks since it adds an element of unpredictability to the input data during inference. Moreover,  the defense helps to improve the overall security of the deep learning classifier by making it more difficult for attackers to generate effective adversarial examples.
\begin{table}[htbp]
\centering
\caption{Optimization-based adversarial attack success rate with and without using the proposed defense mechanism}
\label{tab:ASR-opt}
\begin{tabular}{ccccc}
\hline
Model                                                & FGSM & IGSM & PGD & C\&W \\ \hline
MNIST-No defense                                     & 56.87\% & 71.53\%    & 85.34\% & 92.1\% \\ \hline
\begin{tabular}[c]{@{}c@{}}MNIST-with defense\\ r=2\end{tabular}  &  1.64\%   & 6.3\%     &  7.41\%   &   9.45\% \\ \hline
\end{tabular}
\end{table}
\subsection{Proposed defense mechanism performance under generative attacks}
\textbf{\em{Attacks configuration}} We evaluate the proposed defense using generative adversarial attacks. Generative adversarial attacks are shown to be effective since they are able to produce synthetic data that resemble real data, making it more difficult to distinguish between legitimate and adversarial examples.
In this work, we use the generative adversarial attack proposed by \cite{goodfellow2020generative}. The attack involves a generator and a discriminator. The discriminator is a classifier that is used to distinguish between real and synthetic input.  In this context, the discriminator is the classifier, described in Table \ref{tab:architecture}, that the attacker is trying to fool.  The architecture of the generator is typically designed to be complementary to the discriminator and is often informed by the architecture of the discriminator. Table \ref{tab:gan-architecture} describes the architecture of the generator used in this attack. The generator takes random input and outputs images of the same size as the MNIST images ($28\times28\times1$). \\ 
 we train the GAN using the parameters described in Table \ref{tab:training_gan}. During the generative attack training phase, the generator will produce synthetic images, which will be evaluated by the discriminator. The discriminator provides feedback to the generator, indicating how well the generated images reassemble the true image. The generator will use this feedback to update its parameters  such that it will produce a better synthetic example in the next iteration. This process is repeated until the generator produces synthetic samples that can fool the discriminator.  In this case, the discriminator is the trained victim model. Thus, the discriminator  will not be retrained during the  GAN training phase and will be used only to test the output of the generator. At every testing iteration of the synthetic examples, the discriminator will be fed with a randomly selected  image from $R$. 
 
 We, then, test the generative attack using 1000 adversarial images generated by the GAN's generator.  The testing of the generative adversarial attack  follows the test phase workflow outlined in Section \ref{sec:dtl_sol}.
\begin{table}[htbp]
\centering
\caption{Architecture of generator used in the generative adversarial attack}
\label{tab:gan-architecture}
\begin{tabular}{ll}
\hline
\multicolumn{2}{c}{\em Generator} \\ \hline
Layer             & Parameters       \\ \hline
Dense.Relu        & 7*7*28           \\ \hline
Reshape           & 7x7x28            \\ \hline
Upsampling2D      & 2x2               \\ \hline

Conv2DTranspose.ReLU       & 3x3x32           \\ \hline
Conv2DTranspose.ReLU       & 3x3x32           \\ \hline
Upsampling2D       & 2x2              \\ \hline
Conv2DTranspose.ReLU        & 3x3x64           \\ \hline
Conv2DTranspose.ReLU       & 3x3x64           \\ \hline
Conv2DTranspose.ReLU       & 3x3x1           \\ \hline
\end{tabular}
\end{table}

\begin{table}[htbp]
\centering
\caption{Training parameters of GAN}
\label{tab:training_gan}
\begin{tabular}{ll}
\hline
Parameter           & Value \\ \hline
Optimization method & Adam   \\ \hline
Learning Rate       & 0.01  \\ \hline
Loss function       & categorical crossentropy    \\ \hline
Epochs              & 64 and 128    \\ \hline
Input Noise         & 100    \\ \hline
\end{tabular}
\end{table}

Table \ref{tab:ASR-gen} shows the attack success rate of the  generative adversarial attack  when trained using 64 and 128 epochs, respectively, on the classifier with and without using the defense mechanism.  This process is repeated five times, and the reported success rate is an average of these experiments.
The defense mechanism is evaluated using a subset of two random images ($R=\{R_0, R_1\}$. The results indicate that without any defense mechanism, the success rate of the attacks of the generative adversarial attack is 100\% for both epochs =64 and 128. However, when the defense mechanism is applied,
we observe that the attack success rate dropped to $9.68\%(<10\%)$ and $21.03\%$ when the adversarial model was trained for 64 and 128 epochs, respectively. These results suggest that the defense mechanism is quite effective in reducing the success rate of attacks. Indeed, using only two random projections ($r=2$), the defense mechanism has achieved a significant reduction of $90.32\%$ and $78.97\%$ in the attack success rate, when the adversarial model was trained for 64 and 128 epochs, respectively.

Specifically, the findings reveal that for the adversarial model trained for 64 epochs, the defense mechanism  successfully reduces the attack success rate to less than 10\%, indicating that the attacker's  generator  was not able to produce effective adversarial examples capable of fooling the discriminator. Note that 10\% would be the average success rate for a random choice, as discussed above. 
However, when the generative adversarial model is trained for $128$ epochs, we observe that the attack success rate has surpassed  the random classification rate of $10\%$ and achieved a success rate of $21.03\%$. These findings indicate that the attacker was successful in generating some adversarial examples that can evade the defenses of the target system, but the success rate is still relatively low. Furthermore, the findings suggest that using only a subset of two random images, the proposed mechanism sees its effectiveness degraded when confronted with generative adversarial attacks. The reason is that the attacker is using a generative model to create adversarial examples that can evade the classifier's defenses. 
If the defense framework uses only two random images, the attacker can  learn, after a few iterations, to  generate adversarial examples that are optimized to bypass the countermeasure implemented by these two random images. 
An intuitive solution to this issue would be to use a higher number of random images to strengthen the proposed defenses against generative attacks. The impact of the number of random images used by the proposed defense will be studied in the next subsection.
\begin{table}[htbp]
\centering
\caption{Generative adversarial attack success rate on the defense mechanism}
\label{tab:ASR-gen}
\begin{tabular}{cc}
\hline
{\em Model}                                                & {\em ASR}  \\ \hline
MNIST-No defense, epochs=64 & 100\% \\ \hline
MNIST-No defense, epochs=128 & 100\% \\ \hline

MNIST-With defense, r=2, epochs=64  &  9.68\%   \\ \hline
MNIST-With defense, r=2, epochs=128\  &  21.03\%   \\ \hline
\end{tabular}
\end{table}
\subsection{Varying the number of hyperspace projections}
\begin{figure}[htbp]
\centerline{\includegraphics[width=\linewidth]{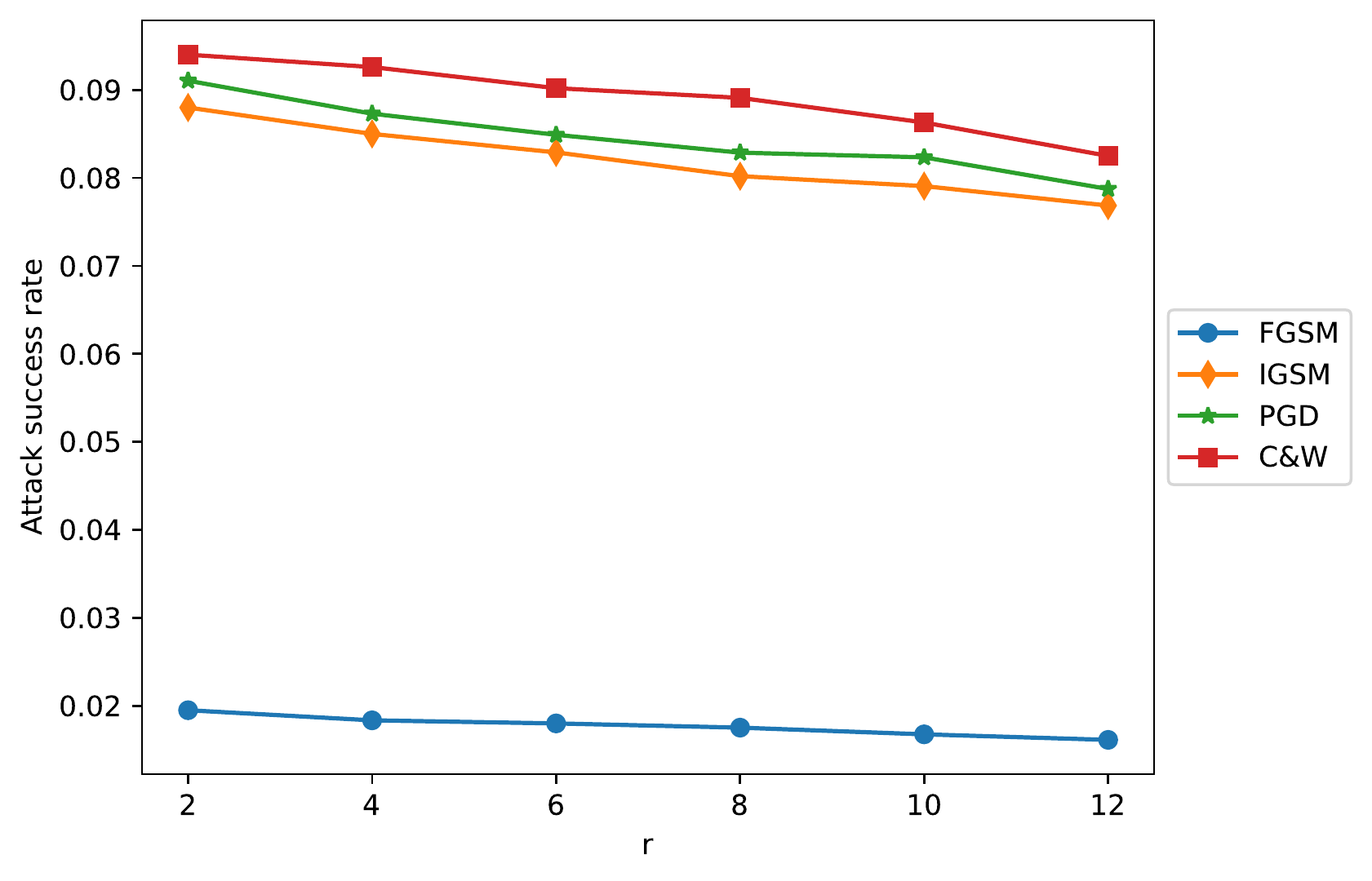}}
\caption{Attack success rate of the proposed defense mechanism under different state-of-the-art optimization-based attacks. }
\label{fig:opt_attack_d}
\end{figure}

  We aim to study the impact of the number of hyperspace projections on the robustness of the proposed defense technique. To this end, we evaluated the proposed defense using different sets of random images $R_i$, with $i=2,4,6,8,10,12$. Each set $R_i$  is created by randomly generating the corresponding  $r=i$ random images as discussed in Section \ref{sec:dtl_sol}. \\
\textbf{\em{Optimization-based adversarial attacks}}\\
The impact of varying the number of random images used on optimization-based attacks is shown in Fig. \ref{fig:opt_attack_d}.
It is  
worth noting that the attack success rate for all the optimization-based attacks is less than $10\%$, suggesting that the defense mechanism is effective in preventing adversarial attacks. FGSM achieves the lowest attack success rate, followed by IGSM, PGD, and C\&W, regardless of the number of random images used. This suggests that FGSM is the least powerful among these attacks, while C\&W is the most powerful. However, all of them show a similar trend of decreasing attack success rate when increasing the number of random images. This is because, at each iteration of the attack, the defense mechanism uses a randomly selected trained classifier---increasing $r$ means that the chances for the adversary to encounter, over two different interactions, a different classifier, increase with $r$. 
Moreover, the randomness of the random images makes it harder for the attacker to find a specific direction in the input space that can fool the defense of the classifier.
Thus, it  becomes increasingly challenging to generate adversarial examples as the number of random images increases.

Finally, we observe that when the number of random images is low, $r=2$or $4$, the state-of-the-art optimization-based attacks can still generate effective adversarial examples similar to the random classification rate For instance, C\&W attack achieves an attack success rate of 9.4\% when the defense mechanism uses a subset of two random images ($r=2$). However, as the number of random images increases, it becomes more difficult for the attacker to create adversarial examples, and the attack success rate decreases. For example, the C\&W attack achieves an attack success rate of 8.25\% when the defense mechanism uses a subset of 12 random images, ($r=12$).
These results support the intuition that the increasing the number of hyperspaces that are utilized---and hence the more trained classifiers are used---the more diverse the decision boundaries of the proposed defense mechanism, and hence the more decision boundaries that the attacker should consider when creating the adversarial examples.

\begin{figure}[htbp]
\centerline{\includegraphics[width=0.8\linewidth]{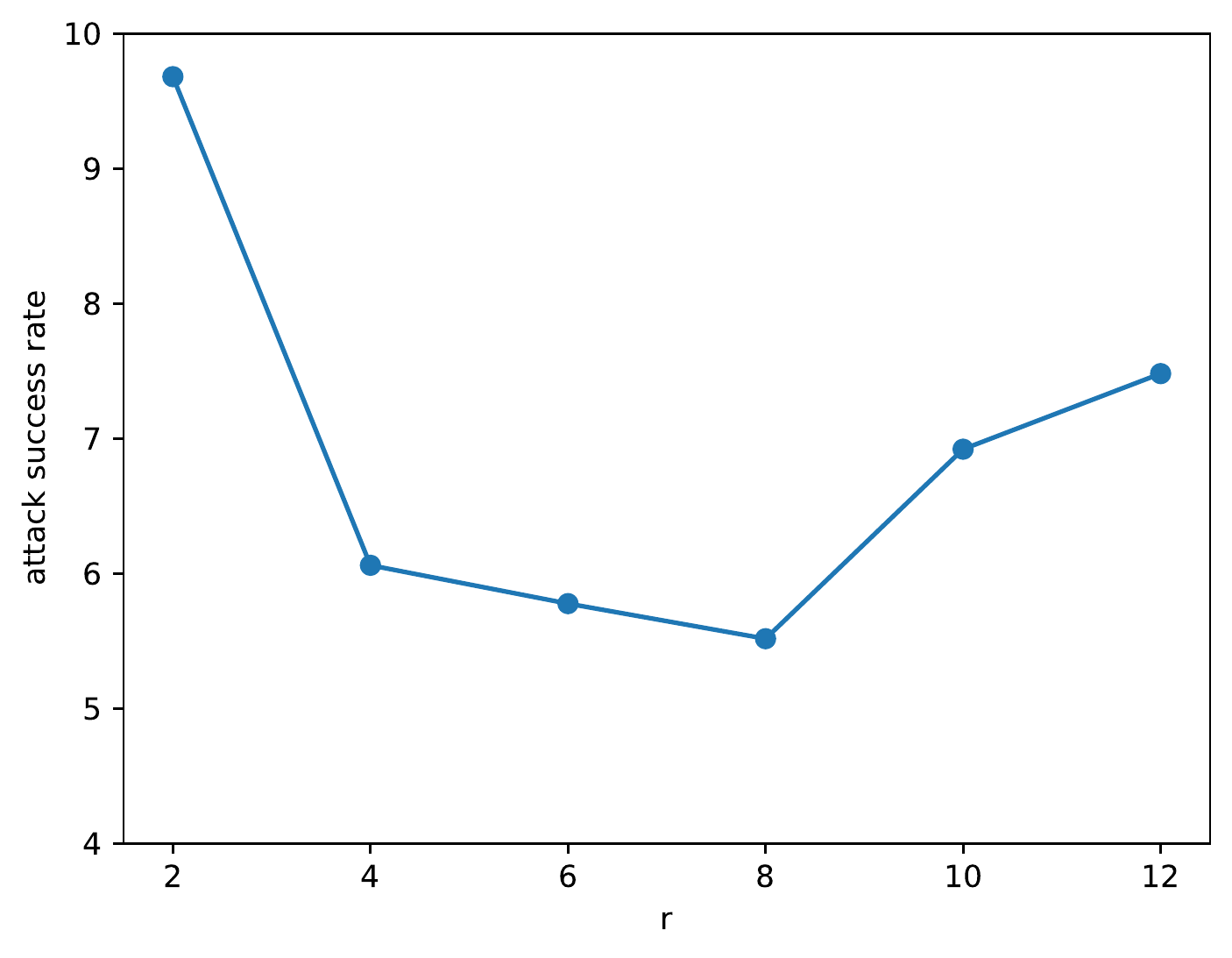}}
\caption{Generative adversarial attack success rate for 64 epochs using different numbers of random images }
\label{fig:ep64}
\end{figure}

 \textbf{\em{Generative adversarial attacks}}
 The results of varying the number of hyperspace used in generative adversarial attacks trained for 64 epochs are shown in Fig. \ref{fig:ep64}. Overall we observe that the attack success rate did not exceed $10\%$. This is a testament to the quality of the defense mechanism that reduces the quality of  a classifier to that of a random choice among the available classes. Indeed, recall that the classifier, for every input, returns a  class among the 10 available ones. Thus, even if the adversarial input is not meaningful or invalid, it still can be classified as one of the 10 labels with a probability of $10\%$.  We also observe that when $r$ increases, the success rate of the adversary decreases, e.g., when $r=2$, the attack success rate is $9.68\%$; however, for $r=4$, the attack success rate is $6.06\%$.
This result can be explained by the fact that generating an adversarial example to fool four models is more complex since it requires combining the perturbation for each individual model. Furthermore, as the attacker tries to optimize the perturbations for more models, it becomes more difficult to find perturbations that can evade the defenses of all the models simultaneously. 
However, we also notice that  the attack success rate increases again when  $r=8$. One possible explanation for this phenomenon is that, as the number of employed hyperspaces increases, the attacker is trying to find the optimal balance between the complexity of the attack and its effectiveness. On the one hand, from the attacker's perspective, the more complex the attack is, the more difficult it is to implement and execute. On the other hand, the more effective the attack is, the higher the probability of successfully evading the defenses of the target system. Therefore, the attacker tries to find the best trade-off between complexity and effectiveness. As the number of hyperspaces increase, the attacker may discover that generating random images or perturbations is more effective than optimizing the adversarial perturbation for all the models. 

\begin{figure}[htbp]
\centerline{\includegraphics[width=0.8\linewidth]{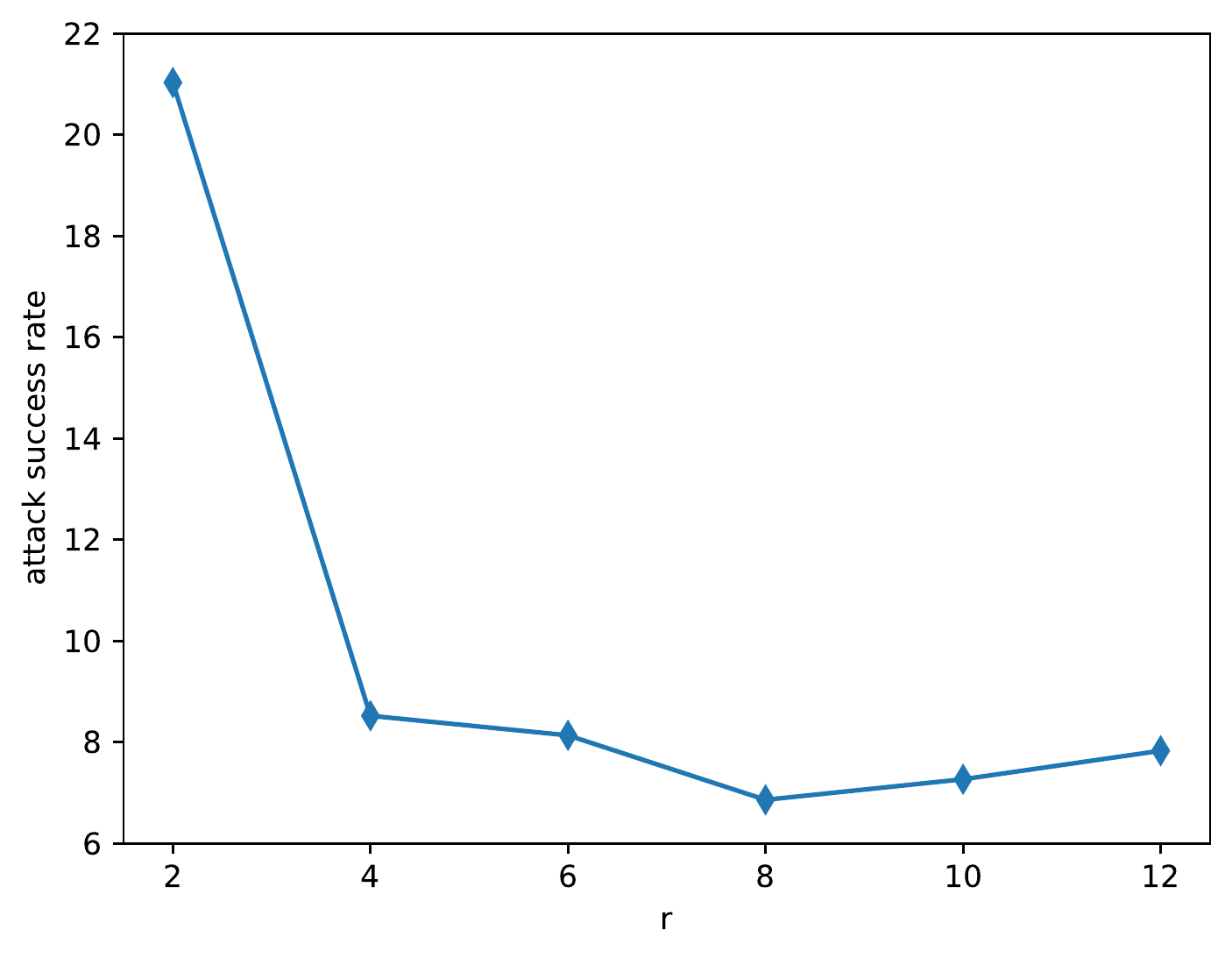}}
\caption{Generative adversarial attack success rate under 128 epochs using  different subsets of random images. }
\label{fig:ep128}
\end{figure}

Fig. \ref{fig:ep128} shows the impact of varying the number of hyperspaces on the generative attack success when the attack is trained for 128 epochs. We observe that the attack success rate  on the proposed defense using a subset of  random images of two, $r=2$, surpassed the random classification rate of $10\%$. This finding suggests that the attacker was able to learn how to generate some adversarial examples that can evade the classifier. 
However, when the number of hyperspaces increases (i.e., $r>2$), the attack success rate gradually decreases. This can be explained by the fact that generating adversarial examples to evade the proposed defense mechanism using multiple hyperspaces is more complex. The generator needs to learn to output an adversarial example that takes into account the interactions between the decision boundaries of each of the trained models obtained by each of the employed hyperspace. 

Interestingly, as the number of hyperspaces is  further increased (i.e., $r>6$), the attack success rate starts to increase again. One possible explanation for this phenomenon can be attributed to the attacker's attempt to find the optimal balance between complexity and effectiveness. Initially, the attacker aims to learn how to generate adversarial examples that can fool the classifier. However, at a certain point, the attacker discovers that generating random images yields better results, prompting it to return to generating random adversarial examples.
Despite the re-increase of the attack success rate, it remains below 10\%, meaning that the proposed defense technique was effective in preventing most adversarial attacks. Finally, we can conclude that increasing the number of hyperspaces---that is, the number of classifiers---can improve the defense against generative adversarial attacks, but it also adds complexity to the defense mechanism. Therefore, finding the optimum number of hyperspaces that balance effectiveness and complexity is an ongoing research that is left for future work.

Overall, our findings demonstrate the effectiveness of using multiple hyperspace projections in the proposed defense mechanism to improve the robustness of the classifier against both optimization-based and generative adversarial attacks. However, there may be a trade-off between the number of random images used, i.e., the number of hyperspace projections, and the computational resources required to train and test the classifier, that will be assessed in future work.

\section{Conclusion}
\label{sec:conclusion}
In this paper, we proposed a novel defense mechanism against adversarial attacks.
We focused on image classification, and hence we used a random  image to project the original dataset into another dataset, which is then used to train a classifier. Using more than one random image (say $r$), 
this process returns $r$ different classifiers with different decision boundaries. 
During the inference step, the defense mechanism randomly projects the inputs using a random image and uses the corresponding classifier---the one built with the dataset projected according to the random image used for testing the input. This increases the diversity of the decision boundaries of the classifier, making it harder for the attacker to find a specific direction in the input space that can fool the classifier while preserving the classification accuracy for legitimate inputs. Indeed, the decrease in the accuracy of classification
is almost negligible---less than  1.2\% on average.
Our experimental results further confirmed the effectiveness of the proposed defense mechanism. In particular, we tested our solution against optimization-based attacks (FGSM, IGSM, PGD, and C\&W) as well as generative attacks using the MNIST dataset.  The results showed that our defense mechanism significantly reduces the attack success rate of both optimization-based and generative adversarial attacks---by at least $89\%$ and 78\%, respectively.
The novel approach, the sound methodology, and the achieved results, other than being valuable on their own, also pave the way for further research along the directions highlighted in the paper. 

\balance
\bibliographystyle{plainnat}
\bibliography{ARXIV.bib}

\begin{thebibliography}{31}
\providecommand{\natexlab}[1]{#1}
\providecommand{\url}[1]{\texttt{#1}}
\expandafter\ifx\csname urlstyle\endcsname\relax
  \providecommand{\doi}[1]{doi: #1}\else
  \providecommand{\doi}{doi: \begingroup \urlstyle{rm}\Url}\fi

\bibitem[Akhtar et~al.(2021)Akhtar, Mian, Kardan, and Shah]{akhtar2021advances}
Naveed Akhtar, Ajmal Mian, Navid Kardan, and Mubarak Shah.
\newblock Advances in adversarial attacks and defenses in computer vision: A
  survey.
\newblock \emph{IEEE Access}, 9:\penalty0 155161--155196, 2021.

\bibitem[Ali et~al.(2022)Ali, Mohammed, and Ahmad]{ali2022evaluating}
Ziad Tariq~Muhammad Ali, Ameer Mohammed, and Imtiaz Ahmad.
\newblock Evaluating adversarial robustness of secret key-based defenses.
\newblock \emph{IEEE Access}, 10:\penalty0 34872--34882, 2022.

\bibitem[AprilPyone and Kiya(2021)]{aprilpyone2021block}
MaungMaung AprilPyone and Hitoshi Kiya.
\newblock Block-wise image transformation with secret key for adversarially
  robust defense.
\newblock \emph{IEEE Transactions on Information Forensics and Security},
  16:\penalty0 2709--2723, 2021.

\bibitem[Carlini and Wagner(2017{\natexlab{a}})]{carlini2017adversarial}
Nicholas Carlini and David Wagner.
\newblock Adversarial examples are not easily detected: Bypassing ten detection
  methods.
\newblock In \emph{ACM workshop on artificial intelligence and security}, pages
  3--14, 2017{\natexlab{a}}.

\bibitem[Carlini and Wagner(2017{\natexlab{b}})]{carlini2017towards}
Nicholas Carlini and David Wagner.
\newblock Towards evaluating the robustness of neural networks.
\newblock In \emph{2017 IEEE S\&P}, pages 39--57. Ieee, 2017{\natexlab{b}}.

\bibitem[Carlini and Wagner(2018)]{carlini2018audio}
Nicholas Carlini and David Wagner.
\newblock Audio adversarial examples: Targeted attacks on speech-to-text.
\newblock In \emph{2018 IEEE security and privacy workshops (SPW)}, pages 1--7.
  IEEE, 2018.

\bibitem[Feinman et~al.(2017)Feinman, Curtin, Shintre, and
  Gardner]{feinman2017detecting}
Reuben Feinman, Ryan~R Curtin, Saurabh Shintre, and Andrew~B Gardner.
\newblock Detecting adversarial samples from artifacts.
\newblock \emph{arXiv preprint arXiv:1703.00410}, 2017.

\bibitem[Gong et~al.(2017)Gong, Wang, and Ku]{gong2017adversarial}
Zhitao Gong, Wenlu Wang, and Wei-Shinn Ku.
\newblock Adversarial and clean data are not twins.
\newblock \emph{arXiv preprint arXiv:1704.04960}, 2017.

\bibitem[Goodfellow et~al.(2020)Goodfellow, Pouget-Abadie, Mirza, Xu,
  Warde-Farley, Ozair, Courville, and Bengio]{goodfellow2020generative}
Ian Goodfellow, Jean Pouget-Abadie, Mehdi Mirza, Bing Xu, David Warde-Farley,
  Sherjil Ozair, Aaron Courville, and Yoshua Bengio.
\newblock Generative adversarial networks.
\newblock \emph{Communications of the ACM}, 63\penalty0 (11):\penalty0
  139--144, 2020.

\bibitem[Goodfellow et~al.(2014)Goodfellow, Shlens, and
  Szegedy]{goodfellow2014explaining}
Ian~J Goodfellow, Jonathon Shlens, and Christian Szegedy.
\newblock Explaining and harnessing adversarial examples.
\newblock \emph{arXiv preprint arXiv:1412.6572}, 2014.

\bibitem[Guo et~al.(2017)Guo, Rana, Cisse, and Van
  Der~Maaten]{guo2017countering}
Chuan Guo, Mayank Rana, Moustapha Cisse, and Laurens Van Der~Maaten.
\newblock Countering adversarial images using input transformations.
\newblock \emph{arXiv preprint arXiv:1711.00117}, 2017.

\bibitem[Hinton et~al.(2015)Hinton, Vinyals, and Dean]{hinton2015distilling}
Geoffrey Hinton, Oriol Vinyals, and Jeff Dean.
\newblock Distilling the knowledge in a neural network.
\newblock \emph{arXiv preprint arXiv:1503.02531}, 2015.

\bibitem[Kang et~al.(2020)Kang, Song, Du, and Guizani]{kang2020adversarial}
Xu~Kang, Bin Song, Xiaojiang Du, and Mohsen Guizani.
\newblock Adversarial attacks for image segmentation on multiple lightweight
  models.
\newblock \emph{IEEE Access}, 8:\penalty0 31359--31370, 2020.

\bibitem[Kurakin et~al.(2018)Kurakin, Goodfellow, and
  Bengio]{kurakin2018adversarial}
Alexey Kurakin, Ian~J Goodfellow, and Samy Bengio.
\newblock Adversarial examples in the physical world.
\newblock In \emph{Artificial intelligence safety and security}, pages 99--112.
  Chapman and Hall/CRC, 2018.

\bibitem[Li and Li(2017)]{li2017adversarial}
Xin Li and Fuxin Li.
\newblock Adversarial examples detection in deep networks with convolutional
  filter statistics.
\newblock In \emph{Proceedings of the IEEE international conference on computer
  vision}, pages 5764--5772, 2017.

\bibitem[Liu et~al.(2018{\natexlab{a}})Liu, Cheng, Zhang, and
  Hsieh]{liu2018towards}
Xuanqing Liu, Minhao Cheng, Huan Zhang, and Cho-Jui Hsieh.
\newblock Towards robust neural networks via random self-ensemble.
\newblock In \emph{Proceedings of the European Conference on Computer Vision
  (ECCV)}, pages 369--385, 2018{\natexlab{a}}.

\bibitem[Liu et~al.(2018{\natexlab{b}})Liu, Li, Wu, and Hsieh]{liu2018adv}
Xuanqing Liu, Yao Li, Chongruo Wu, and Cho-Jui Hsieh.
\newblock Adv-bnn: Improved adversarial defense through robust bayesian neural
  network.
\newblock \emph{arXiv preprint arXiv:1810.01279}, 2018{\natexlab{b}}.

\bibitem[Lu et~al.(2017)Lu, Issaranon, and Forsyth]{lu2017safetynet}
Jiajun Lu, Theerasit Issaranon, and David Forsyth.
\newblock Safetynet: Detecting and rejecting adversarial examples robustly.
\newblock In \emph{Proceedings of the IEEE international conference on computer
  vision}, pages 446--454, 2017.

\bibitem[Madry et~al.(2017)Madry, Makelov, Schmidt, Tsipras, and
  Vladu]{madry2017towards}
Aleksander Madry, Aleksandar Makelov, Ludwig Schmidt, Dimitris Tsipras, and
  Adrian Vladu.
\newblock Towards deep learning models resistant to adversarial attacks.
\newblock \emph{arXiv preprint arXiv:1706.06083}, 2017.

\bibitem[Meng and Chen(2017)]{meng2017magnet}
Dongyu Meng and Hao Chen.
\newblock Magnet: a two-pronged defense against adversarial examples.
\newblock In \emph{ACM SIGSAC CCS}, pages 135--147, 2017.

\bibitem[Metzen et~al.(2017)Metzen, Genewein, Fischer, and
  Bischoff]{metzen2017detecting}
Jan~Hendrik Metzen, Tim Genewein, Volker Fischer, and Bastian Bischoff.
\newblock On detecting adversarial perturbations.
\newblock \emph{arXiv preprint arXiv:1702.04267}, 2017.

\bibitem[Nowroozi et~al.(2022)Nowroozi, Mohammadi, Golmohammadi, Mekdad, Conti,
  and Uluagac]{nowroozi2022resisting}
Ehsan Nowroozi, Mohammadreza Mohammadi, Pargol Golmohammadi, Yassine Mekdad,
  Mauro Conti, and Selcuk Uluagac.
\newblock Resisting deep learning models against adversarial attack
  transferability via feature randomization.
\newblock \emph{arXiv preprint arXiv:2209.04930}, 2022.

\bibitem[Papernot et~al.(2016)Papernot, McDaniel, Wu, Jha, and
  Swami]{papernot2016distillation}
Nicolas Papernot, Patrick McDaniel, Xi~Wu, Somesh Jha, and Ananthram Swami.
\newblock Distillation as a defense to adversarial perturbations against deep
  neural networks.
\newblock In \emph{2016 IEEE S\&P}, pages 582--597. IEEE, 2016.

\bibitem[Raff et~al.(2019)Raff, Sylvester, Forsyth, and
  McLean]{raff2019barrage}
Edward Raff, Jared Sylvester, Steven Forsyth, and Mark McLean.
\newblock Barrage of random transforms for adversarially robust defense.
\newblock In \emph{Proceedings of the IEEE/CVF CVPR}, pages 6528--6537, 2019.

\bibitem[Shaham et~al.(2015)Shaham, Yamada, and
  Negahban]{shaham2015understanding}
Uri Shaham, Yutaro Yamada, and Sahand Negahban.
\newblock Understanding adversarial training: Increasing local stability of
  neural nets through robust optimization.
\newblock \emph{arXiv preprint arXiv:1511.05432}, 2015.

\bibitem[Taran et~al.(2019)Taran, Rezaeifar, Holotyak, and
  Voloshynovskiy]{taran2019defending}
Olga Taran, Shideh Rezaeifar, Taras Holotyak, and Slava Voloshynovskiy.
\newblock Defending against adversarial attacks by randomized diversification.
\newblock In \emph{Proceedings of the IEEE/CVF CVPR}, pages 11226--11233, 2019.

\bibitem[Wang et~al.(2009)Wang, Ma, and Zhou]{wang2009brief}
Hua Wang, Cuiqin Ma, and Lijuan Zhou.
\newblock A brief review of machine learning and its application.
\newblock In \emph{2009 international conference on information engineering and
  computer science}, pages 1--4. IEEE, 2009.

\bibitem[Xie et~al.(2017)Xie, Wang, Zhang, Ren, and Yuille]{xie2017mitigating}
Cihang Xie, Jianyu Wang, Zhishuai Zhang, Zhou Ren, and Alan Yuille.
\newblock Mitigating adversarial effects through randomization.
\newblock \emph{arXiv preprint arXiv:1711.01991}, 2017.

\bibitem[Yann(1998)]{yann1998mnist}
Lecun Yann.
\newblock The mnist database of handwritten digits.
\newblock \emph{R}, 1998.

\bibitem[You et~al.(2019)You, Ye, Li, Xu, and Wang]{you2019adversarial}
Zhonghui You, Jinmian Ye, Kunming Li, Zenglin Xu, and Ping Wang.
\newblock Adversarial noise layer: Regularize neural network by adding noise.
\newblock In \emph{2019 IEEE International Conference on Image Processing
  (ICIP)}, pages 909--913. IEEE, 2019.

\bibitem[Yu et~al.(2022)Yu, Han, Shen, Yu, Gong, Gong, and
  Liu]{yu2022understanding}
Chaojian Yu, Bo~Han, Li~Shen, Jun Yu, Chen Gong, Mingming Gong, and Tongliang
  Liu.
\newblock Understanding robust overfitting of adversarial training and beyond.
\newblock In \emph{International Conference on Machine Learning}, pages
  25595--25610. PMLR, 2022.

\end{thebibliography}

\end{document}